# Comparing Complex Concepts with Transformers

## Matching Patent Claims Against Natural Language Text

Matthias Blume
IP Aptly, Inc.
San Diego, CA, USA
matthias.blume@ipaptly.com

Ghobad Heidari
IP Aptly, Inc.
San Diego, CA, USA
ghobad.heidari@ipaptly.com

Christoph Hewel
IP Aptly, Inc.
Munich, Germany
christoph.hewel@ipaptly.com

## ABSTRACT

A key capability in managing patent applications or a patent portfolio is comparing claims to other text, e.g. a patent specification. Because the language of claims is different from language used elsewhere in the patent application or in non-patent text, this has been challenging for computer based natural language processing. We test two new LLM-based approaches and find that both provide substantially better performance than previously published values. The ability to match dense information from one domain against much more distributed information expressed in a different vocabulary may also be useful beyond the intellectual property space.

## 1 INTRODUCTION

A patent application consists of claims and other text. The claims very densely represent the key aspects of the invention. They are written to be as general as possible and utilize a distinct vocabulary and grammar: each claim is limited to at most one sentence, and that sentence typically does not have a subject-verb-object structure. The remainder of the patent application is similar to technical text from its domain, which again differs from other types of text such as marketing documents.

A patent examiner must search for prior art documents in order to determine whether the claimed invention is novel and may be allowable, or whether all aspects of the claim have previously been disclosed. A patent owner may want to search a database of documents or all text on the web in order to find products that potentially infringe on the inventions, as specified in the claims. An entity defending itself against infringement may attempt to invalidate a patent by finding novelty-destroying prior art to that patent. In all cases, the key task is to search through a set of documents and determine whether those documents cover all aspects of each claim of the subject patent application or granted patent. Thus, a claim of a subject patent (application) may be considered a query to an information retrieval system whose objective is to retrieve a document or set of documents that contain all aspects of that claim.

Risch et al. [2021] identified EPO Search Reports as a potential source of ground truth data for training and evaluating models specific to matching patent claims against prior patent applications. The European Patent Office trained and evaluated Sentence Transformer models on this data. Our approach is similar but different in several important ways.

Section 2 of this paper describes the US and EPO patent application data, the EPO Search Report data, and our parsing of these datasets. Section 3 describes our algorithms for fine-tuning large language models (LLMs) for the purpose of comparing claims against non-claim text (e.g. from patent specifications) and scoring document similarity with respect to a patent claim. Section 4 describes our results and compares them to previous published results. Section 5 describes a proof-of-concept system for using a claim as a query for real-time semantic search of a large corpus of documents. Section 6 provides conclusions.

## 2 DATA

The basis of our dataset is the EPO's *EP full-text data for text analytics*[1] and USPTO's *Patent Application Full Text Data (No Images)*[2] bulk download data sets. The EPO data includes full-text and metadata of all patent applications and patent documents published by the EPO from 1978 through July, 2022. The USPTO data used for this study includes full-text and metadata of all patent applications published by the USPTO from March 15, 2001 through July, 2022.

The EPO data includes *search reports* from 2012 onwards that can be used to create a labeled dataset for supervised training and evaluation. In these reports, patent examiners cite prior art documents that are relevant for judging the novelty of each application claim. We utilize two categories of citations provided by the examiners: an "X" document negates the novelty of the claimed invention, and an "A" document is a relevant prior art that however *does not* negate the novelty or inventive step.

---

[1] https://www.epo.org/en/searching-for-patents/data/bulk-data-sets/text-analytics
[2] https://bulkdata.uspto.gov/

Each citation references the relevant passages of the prior art document, e.g. "`abstract; figure 1; paragraph [0002] - paragraph [0023]; claims 1-13`". We parse and standardize this field. We keep only references to abstract, claims, and paragraphs and discard references to figures, page and line number, and some other rarer formats. Linking the passage references to the full text of EPO and USPTO patent applications yields a dataset with ground truth of not only which document, but *which passages* are prior art to a specific claim.

Our Search Reports dataset includes 467,558 claim 1 "A" and "X" citations. It seems almost ideal for training a classifier that distinguishes between text passages that cover all aspects or do not cover all aspects of a particular claim. However, the data has several limitations and peculiar characteristics. 1) The examiner may initially identify a document as an "X" citation but, after rebuttal by the inventors, allow that it is not novelty-destroying after all. Thus, "X" citations are not really "ground truth". 2) "A" citations are more likely than "X" citations to reference a small amount of text (fewer paragraphs of the prior art document), e.g., 30% vs 19% reference text with a total length of fewer than 3000 characters. 3) "A" citations are more likely than "X" citations to reference EPO applications: 27% vs 24%. Points #2 and #3 make it possible to build a model that distinguishes between "X" and "A" citations but would be useless in practice.

Both EPO and USPTO (since 2001) delimit different claim elements via <claim-text> XML tags, e.g.:

```
<claim id="CLM-00001">
<claim-text>. A hip protecting device for inflating a poc
ket over a hip joint of a wearer of the device upon a fal
l comprising:
<claim-text>a belt; </claim-text>
<claim-text>a substantially gas impermeable first pocket
fixedly suspended … from said belt; </claim-text>
…
</claim-text>
</claim>
```

This tag is used to split claims into elements (Section 3.2.2).

## 3 METHODS

### 3.1 Model Training

We split the search reports into 80% training and 20% test/evaluation sets by subject patent application ID. That is, each query EPO patent application occurs only in the training set or only in the test set and not both.

For each "X" and "A" citation in the search reports, we concatenate the text of all cited paragraphs, claims, abstracts, and figure descriptions. We split each cited text into chunks of maximum length MaxSeqLength, respecting paragraph boundaries. That is, the beginning of a chunk is always aligned with the beginning of a paragraph. The context window size of our base model defines *MaxSeqLength = 512 tokens*. For each claim 1, we choose pairs of "X" and "A" chunks, using each chunk at most once. For example, if there are five "X" chunks and three "A" chunks, we create three records, each with one "X" and one "A" chunk. This yields 171,323 training records where the "X" chunk is more relevant to the claim than the "A" chunk. We create a second set of records where each of the "A" chunks is used as the positive example and a random "X" chunk is used as the negative example. I.e., the negative example in this second set is an "X" citation for a *different* claim 1. This prevents the model from learning that some chunks are inherently positive or negative due to overall differences between "X" and "A" citations chosen by the examiners.

We use contrastive learning to tune a model such that the similarity between a relevant chunk and the query claim 1 is greater than the similarity between the less relevant chunk and the query claim 1. Specifically, we use the Sentence Transformers technique [5] to fine tune the distilroberta-base[3] model. (This "small LLM" will not yield the highest possible performance. Rather, we chose it for rapid training and evaluation of the techniques described in the next section.) Below, we refer to this fine-tuned model as "CCX", short for claim-chunk transformer.

### 3.2 Similarity Measurement

Given the models described in the previous section, we can compare text from a patent claim against arbitrary text from a different patent application.

The PatentMatch [3] and SearchFormer [6] papers describe how to distinguish between a *paragraph* from an "X" document and a paragraph from a different document. But operating at the paragraph level entails several deficiencies. First, a single paragraph does not generally include all aspects of a patent claim and should not be considered an "X" paragraph. Rather, the complete set of paragraphs identified in the Search Report constitute the "X" citation. Second, the examiners' passage identification is less accurate than the X/A classification. For example, an examiner may cite "paragraphs 1-20" for convenience even if some of the paragraphs in the range did not convey key features of the prior art. Finally, the examiner is initially interested in identifying *documents* that contain the prior art, so the scores for multiple paragraphs should be combined to rank documents. (Once a candidate document has been identified, it is desirable to identify those paragraphs of the candidate document which actually anticipate the claim elements.)

Here, we describe two ways to make multiple comparisons between text from a patent claim and text from a different document and then aggregate the results from the multiple comparisons into a single score that can be used to rank documents by their similarity with a patent claim.

*3.2.1 Maximum Chunk-Claim Similarity.* We break each section of a patent application into chunks of maximum length MaxSeqLength at paragraph boundaries. Thus, a chunk will not span multiple sections ("Abstract", "CrossRef", "Background", "Summary", "BriefFig", "Description", "Claims", "Admin") but typically contains multiple paragraphs. We compute the cosine similarity between the query claim 1 (if the full claim is longer than MaxSeqLength, we use only the last elements of the claim) and each target document chunk. The score of the document with respect to the claim is the maximum similarity of any chunk from the document.

---

[3] https://huggingface.co/distilbert/distilroberta-base

*3.2.2 Weighted Sum of Paragraph-Element Similarity.* We split the query claim 1 into multiple claim elements using the `<claim-text>` XML tag as shown above. We compute the similarity between each query claim element and each target document paragraph. We compute the score of the document *with respect to the claim element* and then the score of the document *with respect to the claim* via functions of the cosine similarity between the claim element and paragraph as well as element and paragraph salience characteristics.

## 3.3 GPT 4o

We explored whether GPT 4o can distinguish between "*X*" and "*A*" citations w.r.t. a query claim. We tested the OpenAI API (which relies exclusively on OpenAI's data) and chatgpt.com with file uploads[4], uploading the full text of the *X* and *A* reference documents. The GPT prompts were structured as:

```
The file US20080295019A1.txt contains the text of patent
application US20080295019. The file US20050060664A1.txt
contains the text of patent application US20050060664.
Each of the following lines is an element of a patent
claim. Which patent application better covers all of the
elements, US20080295019 or US20050060664? Choose one or
the other, do not say "neither". Output only
"US20080295019" or "US20050060664".
[query claim element 1]
[query claim element 2]
…
```

The first two sentences above were not used with the API.

## 4 RESULTS

SearchFormer [6] defined a "hard" task of distinguishing between "*X*" and "*A*" cited documents with respect to claim 1 of a patent application and an "easy" task of distinguishing between "*X*"-cited and random documents. Table 1 presents the results of several models and techniques on these two tasks.

| Method | X/A | X/Random |
|---|---|---|
| *PatentMatch 2021* | *54%* | |
| *SearchFormer 2023* | *53.85%* | *98.04%* |
| *IP Rally 2021* | *58%* | |
| Max Chunk-Claim GP BERT | 53.89% | |
| Max Chunk-Claim CCX | 63.05% | 99.61% |
| Weighted Paragraph-Element CCX | 60.46% | |
| GPT 4o internal data only | 52.75% | |
| GPT 4o upload full text | 59.17% | |

**Table 1:** Accuracy by negative example type for several models. Random selection would yield 50% accuracy (binary choice).

The first three rows of Table 1 show PatentMatch "balanced" [3], SearchFormer [6], and IP Rally [2] published performance numbers. Each of these used a different evaluation set, so the numbers should not be compared exactly. A likely explanation for PatentMatch's and SearchFormer's low values is that they evaluated the ability to distinguish between "*X*" and "*A*" *paragraphs* rather than "*X*" and "*A*" documents.

The next three rows compare our two different models and two different aggregation techniques, as described below. Evaluation is on a hold-out set of 20,012 records that was not used for CCX training. Each record contains the query claim 1, one "*X*" citation, and one "*A*" citation. Note that for training the model, as described in Section 3.1, we used one record per matched pair of *X*, *A* chunks from the search reports, whereas for evaluation, we have one record per matched pair of *X*, *A* documents.

*GP BERT* uses the output of the first token (the [CLS] token) of Google's *BERT for Patents* model[5] without further fine-tuning. This model performed poorly in this task, which concurs with the central finding of the Sentence Transformers paper [5]: large language models trained to generate text do not inherently know which type of similarity is relevant for a particular task. E.g., a base LLM could reasonably find any pair of phrases "Method for performing XYZ comprising the following steps" similar because 7 words match exactly. Search reports provide excellent data for supervised training to fine-tune models to focus on the distinctions relevant in the patent domain.

*CCX* is the model described in Section 3.1. Max Chunk-Claim indicates that the *maximum chunk-claim similarity* technique was used to compute the similarity between a document and the query claim 1. Weighted Paragraph-Element indicates that the *weighted sum of paragraph-element similarity* technique was used to compute the similarity between a document and the query claim 1. Both aggregation techniques yield >60% accuracy on the "hard" task: substantially better than previously published values. The *maximum chunk-claim similarity* technique yields 99.61% accuracy on the "easy" task: substantially better than SearchFormer's published value. We did not evaluate the *weighted sum of paragraph-element similarity* aggregation technique on the "easy" task.

Zero shot performance of GPT 4o via the API is poor due to the limited patent text data accessible to the model. Even when uploading the full text of the X and A documents, GPT 4o's performance is only comparable to our much smaller CCX model tuned for claim-chunk similarity. Qin 2024 [4] note that generative LLMs are sensitive to the order of the items in pairwise comparison. Without file uploads, GPT 4o responded with *the document ID that appeared first in the prompt* 65.5% of the time. GPT 4o's explanations of which patent application better covers the query claim are well-phrased and compelling even when the conclusion disagrees with the EPO search reports. Risch et al. [2021] stated: "The complex linguistic patterns, the legal jargon, and the patent-domain-specific language make it sheer impossible for laymen to manually solve this task."

---

[4] https://help.openai.com/en/articles/8555545-file-uploads-faq

[5] https://huggingface.co/anferico/bert-for-patents

Vowinckel and Hähnke [2023] "found that hard negatives (*A*-citations) alone are too challenging". Kallio [2021] stated "It doesn't tell much about the search results directly, as the *A* citations are good and important parts too." We concur that it is a difficult task and that the *A* citations are also relevant. Our results demonstrate that it *is* possible to achieve far better than random performance, and we anticipate further improvements by using a better base model fine tuned with more data.

A major question was whether an LLM can effectively represent a set of concepts as complex as a patent claim in its hidden layer, and whether comparison of these vector embeddings could be effective for comparing similar concepts. The Max Chunk-Claim approach relies only on this embedding, and the performance indicates that comparing embedding vectors can in fact identify similarity of concepts as complex as a patent claim. The Weighted Paragraph-Element approach breaks down the complex concept into several smaller snippets and compares at a paragraph level rather than a larger chunk of text. Initial results do not demonstrate a substantial improvement over letting the LLM do all the work. A few possible reasons for this are:

1) The weighting scheme is arbitrary (an optimal weighting scheme could be learned from the data).

2) CCX was tuned to compare the similarity of claims and chunks, not elements and paragraphs. A model trained for the latter scenario should perform better.

## 5 REAL-TIME SEARCH

Above, we described several ways to compare a claim against *one* document using semantic vector embeddings of a LLM. Using approximate nearest neighbors search, it is practical to compare a claim against a corpus of documents, for example *all* published patent applications, in real time. We implemented such a system and describe it at a high level below as a proof of concept. Figure 1 shows the user interface.

We pre-compute one vector per chunk of each document in the corpus using the algorithm described in Section 3.2.1 and store these in a vector database. Vowinckel and Hähnke [2023] state: "There are around 70 million simple patent families with at least one document that contains English text. If the average of 126 paragraphs per document holds true, this corresponds to more than 8.8 billion passages that need to be vectorized." Since we compute a vector per chunk rather than per paragraph, we need only about 20 vectors per patent application rather than 126, or 1.4 billion vectors in total. Using a model with a larger context window would reduce the number of vectors per patent.

Our current vector database represents 3.5 million patent applications and comprises approximately 70 million vectors. Computing the embedding vector for the query and retrieving the ranked list of the nearest 5000 chunks in the vector database takes a small fraction of a second. We further support on-the-fly calculation of an embedding vector for each paragraph in the top *N* retrieved documents and re-ranking based on weighted paragraph-element similarity. Single-threaded on an RTX 4090 GPU, this operation takes less than a minute. Thus, real-time

LLM claim search of a corpus of *all* patent applications and re-ranking the top results is practical.

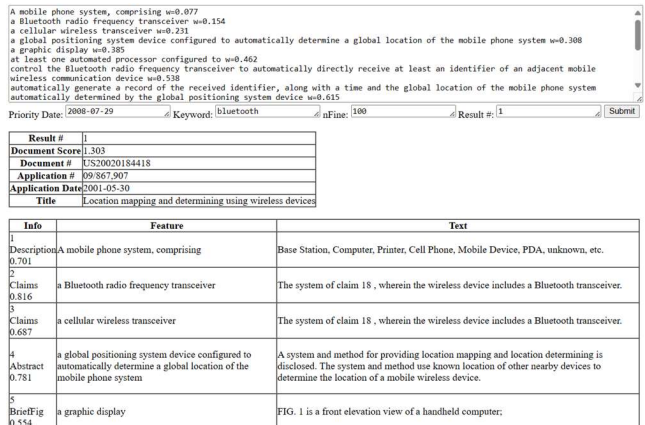

**Figure 1**. IP Aptly real-time patent search user interface.

## 6 CONCLUSIONS

The key features of an invention are densely stated in a patent claim with a peculiar vocabulary and grammar. We demonstrate that it is practical to fine-tune an LLM to compare a claim against a much larger chunk of natural language text. Each chunk should be much larger than a paragraph, as describing the concepts from a single claim typically requires multiple paragraphs of plain text. The "small LLM" fine-tuned for this task performs as well as GPT 4o. To the best of our knowledge, the values published here are the current state of the art on the task of distinguishing between "*X*" and "*A*" citations w.r.t. a query claim.